\documentclass{llncs}
\usepackage{graphicx}
\usepackage{amsfonts}
\usepackage{amsmath}
\usepackage{mathtools}
\usepackage{color}
\usepackage{booktabs}
\usepackage{hyperref}
\usepackage{float,subfig,epstopdf}
\usepackage[capitalise]{cleveref}
\graphicspath{{figures/}}

\definecolor{mypink1}{rgb}{0.858, 0.188, 0.478}
\definecolor{deepmagenta}{rgb}{0.8, 0.0, 0.8}

\usepackage{soul}

\hypersetup{
    colorlinks=true,
    linkcolor=black,
    citecolor=black,
    filecolor=black,
    urlcolor=black,
}

\author{Sofia Ira Ktena\thanks{The support of the EPSRC CDT (HiPEDS, Grant Reference EP/L016796/1) is greatfully acknowledged. Email correspondence to: ira.ktena@imperial.ac.uk} \and Sarah Parisot \and Enzo Ferrante \and Martin Rajchl \and Matthew Lee \and Ben Glocker  \and Daniel Rueckert}
\authorrunning{Ktena et al.}   
\tocauthor{author1 (Affiliation of author1),
author2 (Affiliation of Author2), Daniel Rueckert (Affiliation of
Author3), }

\institute{Biomedical Image Analysis Group, Imperial College London, UK 
}

\title{Distance Metric Learning using Graph Convolutional Networks: Application to Functional Brain Networks}

\begin{document}
\maketitle

\begin{abstract}
Evaluating similarity between graphs is of major importance in several computer vision and pattern recognition problems, where graph representations are often used to model objects or interactions between elements. The choice of a distance or similarity metric is, however, not trivial and can be highly dependent on the application at hand. In this work, we propose a novel metric learning method to evaluate distance between graphs that leverages the power of convolutional neural networks, while exploiting concepts from spectral graph theory to allow these operations on irregular graphs. We demonstrate the potential of our method in the field of connectomics, where neuronal pathways or functional connections between brain regions are commonly modelled as graphs. In this problem, the definition of an appropriate graph similarity function is critical to unveil patterns of disruptions associated with certain brain disorders. Experimental results on the ABIDE dataset show that our method can learn a graph similarity metric tailored for a clinical application, improving the performance of a simple $k$-nn classifier by 11.9\% compared to a traditional distance metric.
\end{abstract}

\section{Introduction}
The highly challenging problem of inexact graph matching entails the evaluation of how much two graphs share or, conversely, how much they differ~\cite{livi2013graph}. Obtaining a measure of global similarity between two graphs can facilitate classification and clustering problems. This concept is particularly valuable in brain connectivity studies, which involve the representation of the structural and/or functional connections within the brain as labelled graphs. Resting-state fMRI (rs-fMRI) can be used to map the connections between spatially remote regions in order to obtain functional networks incorporating the strength of these connections in their edge labels. At the same time, disruptions to this functional network organisation have been associated with neurodevelopmental disorders, such as autism spectrum disorder (ASD)~\cite{abraham2016deriving}. As a result, studying the brain's organisation has the potential to identify predictive biomarkers for neurodevelopmental disorders, a task of great importance for understanding the disorder's underlying mechanisms. Such tasks require an accurate metric of similarity/distance between brain networks to apply statistical and machine learning analyses.

\noindent\textbf{Related work:} 
The estimation of (dis)similarity between two graphs has, most commonly, been dealt with using four mainstream approaches~\cite{livi2013graph}: graph kernels, graph embedding, motif counting and graph edit distance. Graph kernels have been employed to compare functional brain graphs~\cite{takerkart2014graph}, but often fail to capture global properties as they compare features of smaller subgraphs. Graph embedding involves obtaining a feature vector representation that summarizes the graph topology in terms of well-known network features. This method has been widely used to estimate brain graph similarity~\cite{abraham2016deriving}, since it facilitates the application of traditional classification or regression analyses. However, it often discards valuable information about the graph structure. Counting motifs, \emph{i.e.} occurrences of significant subgraph patterns, has also been used~\cite{shervashidze2009efficient}, but is a computationally expensive process. Finally, methods based on graph edit distance neatly model both structural and semantic variation within the graphs and are particularly useful in cases of unknown node correspondences~\cite{raj2010network}, but are limited by the fact that they require the definition of the edit costs in advance.

Recently, different neural network models have been explored to learn a similarity function that compares images patches \cite{zagoruyko2015learning,kumar2016learning}. The network architectures investigated employ 2D convolutions to yield hierarchies of features and deal with the different factors that affect the final appearance of an image. However, the application of convolutions on irregular graphs, such as brain connectivity graphs, is not straightforward. One of the main challenges is the definition of a local neighbourhood structure, which is required for convolution operations. Recent work has attempted to address this challenge by employing a graph labelling procedure for the construction of a receptive field~\cite{niepert2016learning}, but requires node features to meet certain criteria dictated by the labelling function (\emph{e.g.} categorical values). Shuman et al. ~\cite{shuman2013emerging} introduced the concept of signal processing on graphs, through the use of computational harmonic analysis to perform data processing tasks, like filtering. This allows convolutions to be dealt as multiplications in the graph spectral domain, rendering the extension of CNNs to irregular graphs feasible. Recent work by~\cite{defferrard2016convolutional,kipf2016semi} relies on this property to define polynomial filters that are strictly localised and employ a recursive formulation in terms of Chebyshev polynomials that allows fast filtering operations.

\noindent \textbf{Contributions:} In this work, we propose a novel method for learning a similarity metric between irregular graphs with known node correspondences. We use a siamese graph convolutional neural network applied to irregular graphs using the polynomial filters formulated in~\cite{defferrard2016convolutional}. We employ a global loss function that, according to~\cite{kumar2016learning}, is robust to outliers and provides better regularisation. Along with that the network learns latent representations of the graphs that are more discriminative for the application at hand. As a proof of concept, we demonstrate the model performance on the functional connectivity graphs of 871 subjects from the challenging Autism Brain Imaging Data Exchange (ABIDE) database~\cite{di2014autism}, which contains heterogeneous rs-fMRI data acquired at multiple international sites with different protocols. To the best of our knowledge, this is the first application of graph convolutional networks for distance metric learning.

\section{Methodology}
\begin{figure}[t]
\centering
\includegraphics[width=.9\linewidth]{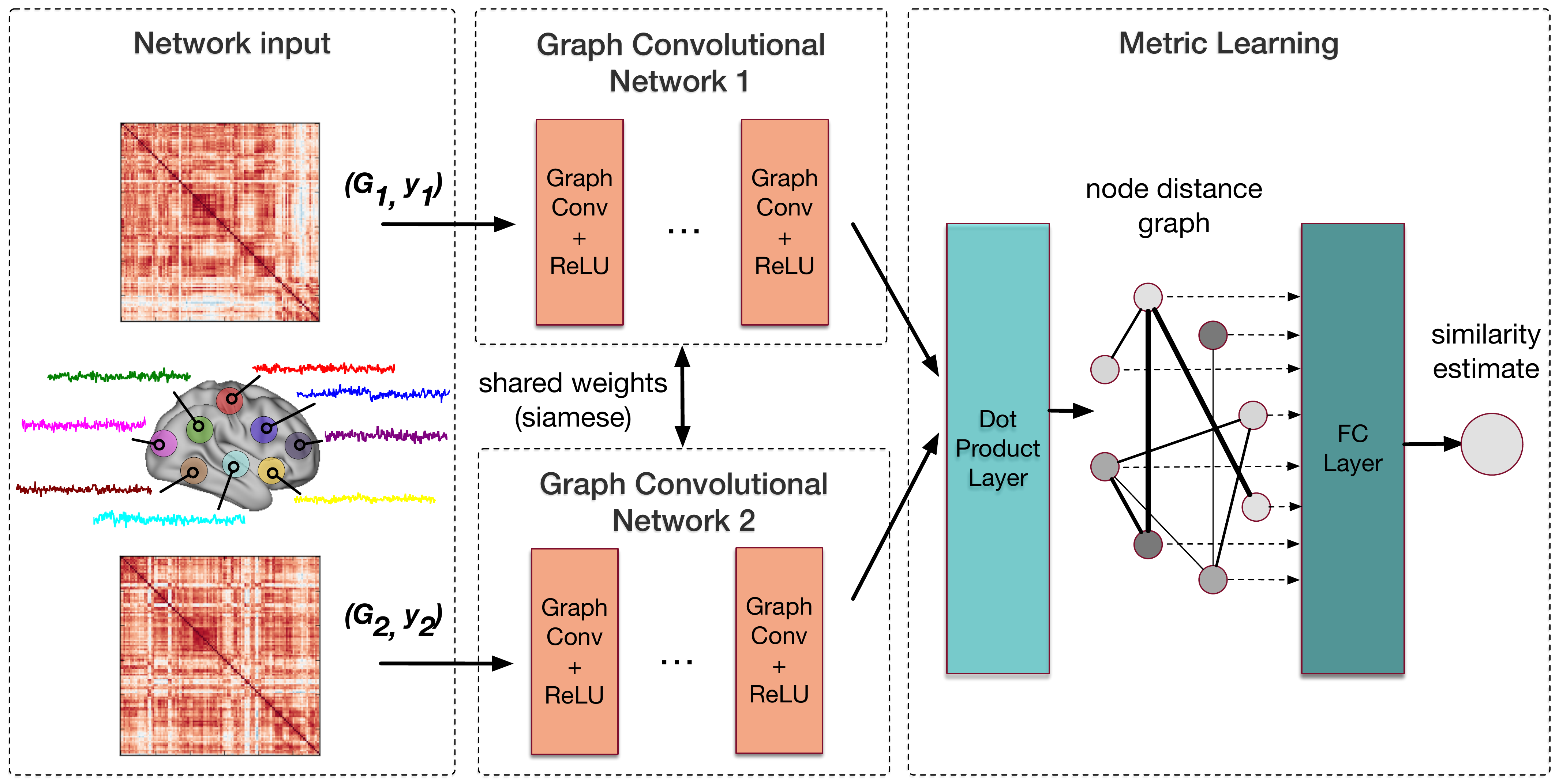}
\caption{Pipeline used for learning to compare functional brain graphs (source code available at \url{https://github.com/sk1712/gcn_metric_learning}).}
\label{fig:overview}
\end{figure}

Fig.~\ref{fig:overview} gives an overview of the proposed model for learning to compare brain graphs. In this section, we first introduce the concept of graph convolutions and filtering in the graph spectral domain in~\ref{subsec:filtering}, as well as the proposed network model and the loss function that we intend to minimise in~\ref{subsec:loss}. Finally, we present the dataset used and the process through which functional brain graphs are derived from fMRI data in~\ref{subsec:dataset}.

\subsection{Spectral Graph Filtering and Convolutions}
\label{subsec:filtering}
The classical definition of a convolution operation cannot be easily generalised to the graph setting, since traditional convolutional operators are only defined for regular grids, e.g. 2D or 3D images.
Spectral graph theory makes this generalisation feasible by defining filters in the graph spectral domain. An essential operator in spectral graph analysis is the normalised graph Laplacian~\cite{shuman2013emerging}, defined as $L = I_R - D^{-1/2} A D^{-1/2}$, where $A \in \mathbb{R}^{R \times R}$ is the adjacency matrix associated with the graph $\mathcal{G}$, $D$ is the diagonal degree matrix and $I_R$ is the identity matrix. $L$ can be decomposed as $L=U \Lambda U^T$, where $U$ is the matrix of eigenvectors and $\Lambda$ the diagonal matrix of eigenvalues. The eigenvalues represent the frequencies of their associated eigenvectors, \emph{i.e.} eigenvectors associated with larger eigenvalues oscillate more rapidly between connected nodes. The graph Fourier transform of a signal $\mathbf{c}$ can, then, be expressed as $\mathbf{c} = U \hat{\mathbf{c}}$. This allows to define a convolution on a graph as a multiplication in the spectral domain of the signal $\mathbf{c}$ with a filter $g_{\theta}=diag(\theta)$ as:

\begin{equation}
g_{\theta} \ast \mathbf{c} = U g_{\theta} U^T \mathbf{c},
\label{eq:convolution}
\end{equation}

\noindent where $\theta \in \mathbb{R}^{R}$ is a vector of Fourier coefficients and $g_{\theta}$ can be regarded as a function of the eigenvalues of $L$, i.e. $g_{\theta}(\Lambda)$~\cite{kipf2016semi}.

To render the filters $K$-localised in space and reduce their computational complexity they can be approximated by a truncated expansion in terms of Chebyshev polynomials of order $K$~\cite{hammond2011wavelets}. The Chebyshev polynomials are recursively defined as $T_k(c)=2cT_{k-1}(c) - T_{k-2}(c)$, with $T_0(c)=1$ and $T_1(c)=c$. The filtering operation of a signal $\mathbf{c}$ with a $K$-localised filter is, then, given by:

\begin{equation}
y = g_{\theta} (L) \mathbf{c} = \sum_{k=0}^{K} \theta_k T_k(\tilde{L}) \mathbf{c},
\end{equation}

\noindent with $\tilde{L}=\frac{2}{\lambda_{max}} L - I_R$, where $\lambda_{max}$ denotes the largest eigenvalue of $L$. The $j^{th}$ output feature map of sample $s$ in a Graph Convolutional Network (GCN) is then given by:

\begin{equation}
y_{s,j} = \sum_{i=1}^{F_{in}} g_{\theta_{i,j}} (L) c_{s,i} \in \mathbb{R}^R,
\end{equation}

\noindent yielding $F_{in} \times F_{out}$ vectors of trainable Chebyshev coefficients $\theta_{i,j} \in \mathbb{R}^K$, where $c_{s,i}$ denotes the input feature maps.

\subsection{Loss Function and Network Architecture}
\label{subsec:loss}
Our siamese network, presented in Fig.~\ref{fig:overview}, consists of two identical sets of convolutional layers sharing the same weights, each taking a graph as input. An inner product layer combines the outputs from the two branches of the network and is followed by a single fully connected (FC) output layer with a sigmoid activation function and one output, that corresponds to the similarity estimate. The FC layer accounts for integrating global information about graph similarity from the preceding localised filters. Each convolutional layer is succeeded by a non-linear activation, i.e. Rectified Linear Unit (ReLU).

We train the network using the pairwise similarity global loss function proposed in~\cite{kumar2016learning} that yields superior results in the problem of learning local image descriptors compared to traditional losses. This loss maximises the mean similarity $\mu^+$ between embeddings belonging to the same class, minimises the mean similarity between embeddings belonging to different classes $\mu^-$ and, at the same time, minimises the variance of pairwise similarities for both matching $\sigma^{2+}$ and non-matching $\sigma^{2-}$ pairs of graphs. The formula of this loss function is given by:

\begin{equation}
J^g = (\ \sigma^{2+} + \sigma^{2-}) \, + \, \lambda \, \max \, (0, m - (\mu^+ - \mu^-)),
\end{equation}

\noindent where $\lambda$ balances the importance of the mean and variance terms, and $m$ is the margin between the means of matching and non-matching similarity distributions. An additional $l_2$ regularisation term on the weights of the fully connected layer is introduced to the loss function.

\subsection{From fMRI Data to Graph Signals}
\label{subsec:dataset}
The dataset is provided by the Autism Brain Imaging Data Exchange (ABIDE) initiative \cite{di2014autism} and has been preprocessed by the Configurable Pipeline for the Analysis of Connectomes (C-PAC) \cite{craddock2013}, which involves skull striping, slice timing correction, motion correction, global mean intensity normalisation, nuisance signal regression, band-pass filtering (0.01-0.1Hz) and registration of fMRI images to standard anatomical space (MNI152). It includes $N=871$ subjects from different imaging sites that met the imaging quality and phenotypic information criteria, consisting of 403 individuals suffering from ASD and 468 healthy controls. We, subsequently, extract the mean time series for a set of regions from the Harvard Oxford (HO) atlas comprising $R=110$ cortical and subcortical ROIs~\cite{desikan2006automated} and normalise them to zero mean and unit variance. 

Spectral graph convolutional networks filter signals defined on a common graph structure for all samples, since these operations are parametrised on the graph Laplacian. As a result, we model the graph structure solely from anatomy, as the $k$-NN graph $\mathcal{G}=\{\mathcal{V},\mathcal{E}\}$, where each ROI is represented by a node $v_i \in \mathcal{V}$ (located at the centre of the ROI) and the edges $\mathcal{E}=\{e_{ij}\}$ of the graph represent the spatial distances between connected nodes using $e_{ij}=d(v_i, v_j)=\sqrt[]{||v_i-v_j||^2}$. For each subject, node $v_i$ is associated with a signal $c_{si} : v_i \rightarrow \mathbb{R}^{R}$, $s = 1, ..., N$ which contains the node's connectivity profile in terms of Pearson's correlation between the representative rs-fMRI time series of each ROI.

\section{Results}

We evaluate the performance of the proposed model for similarity metric learning on the ABIDE database. Similarly to the experimental setup used in~\cite{zagoruyko2015learning}, we train the network on matching and non-matching pairs. In this context, matching pairs correspond to brain graphs representing individuals of the same class (ASD or controls), while non-matching pairs correspond to brain graphs representing subjects from different classes. Although the ground truth labels are binary, the network output is a continuous value, hence training is performed in a weakly supervised setting. To deal with this task, we train a siamese network with 2 convolutional layers consisting of 64 features each. A binary feature is introduced at the FC layer indicating whether the subjects within the pair were scanned at the same site or not. The different network parameters are optimised using cross-validation. We use dropout ratio of 0.2 at the FC layer, regularisation 0.005, learning rate 0.001 with an Adam optimisation and $K=3$, where the filters at each convolution are taking into account neighbours that are at most $K$ steps away from a node. For the global loss function, the margin $m$ is set to 0.6, while the weight $\lambda$ is 0.35. We train the model for 100 epochs on 43000 pairs in mini-batches of 200 pairs. These pairs result from 720 different subjects (after random splitting), comprising 21802 matching and 21398 non-matching graph pairs, and we make sure that all graphs are fed to the network the same number of times to avoid biases. The test set consists of all combinations between the remaining 151 subjects, i.e. 11325 pairs, 5631 of which belong to the same class (either ASD or controls) and 5694 belong to different classes. We also ensure that subjects from all 20 sites are included in both training and test sets.

\begin{figure}[t]
\centering
\subfloat{\includegraphics[width=0.25\textwidth]{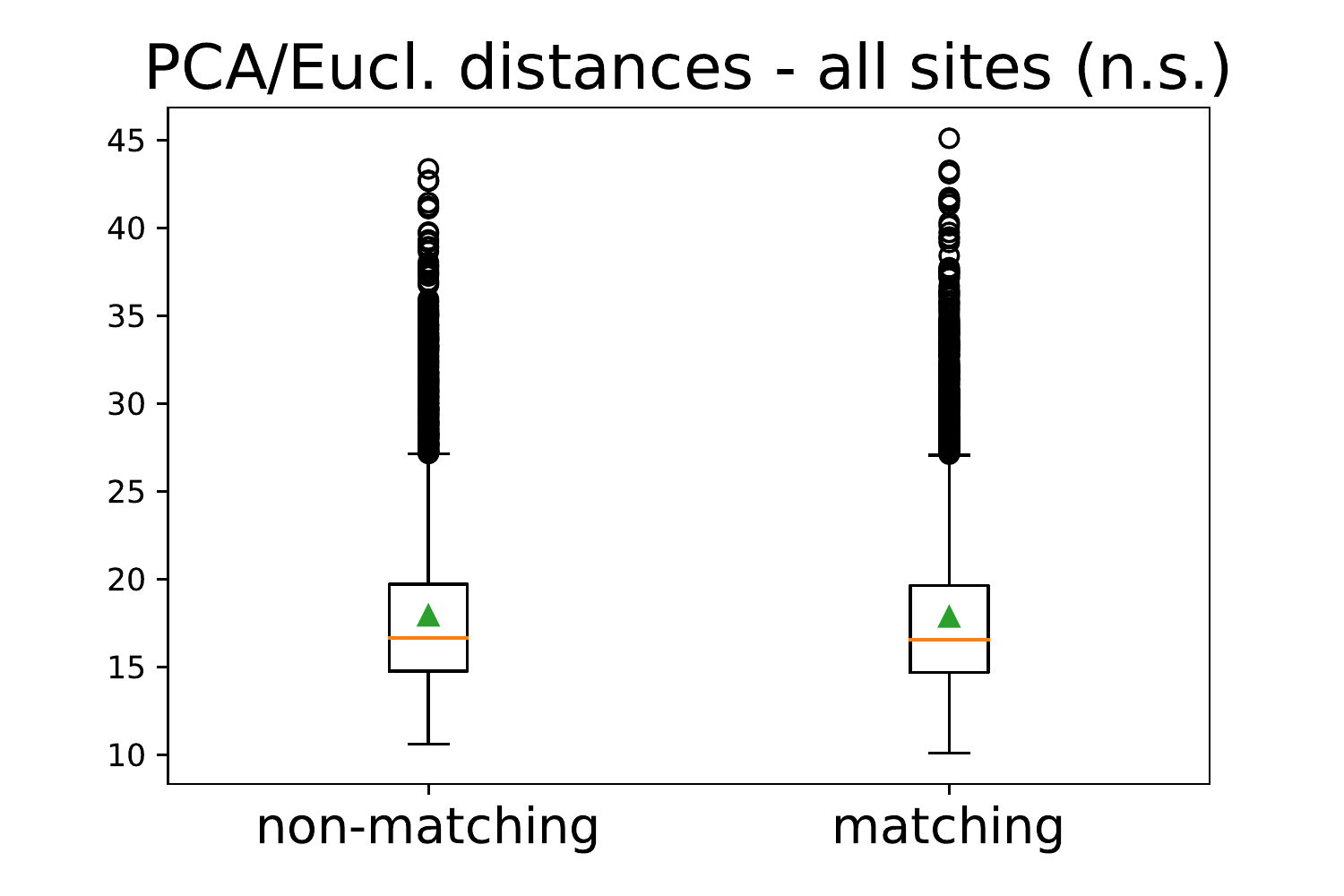}}
\subfloat{\includegraphics[width=0.25\textwidth]{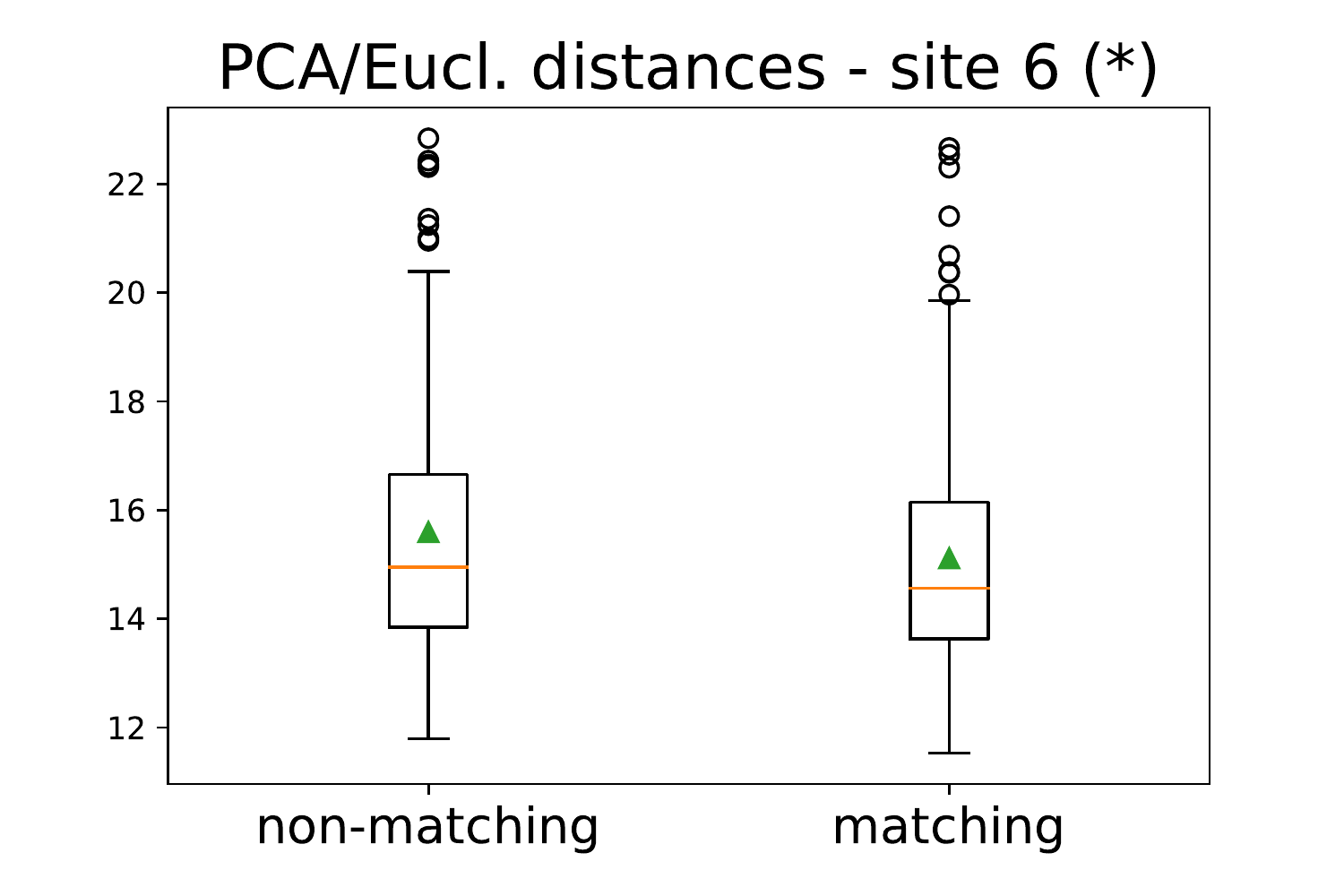}}
\subfloat{\includegraphics[width=0.25\textwidth]{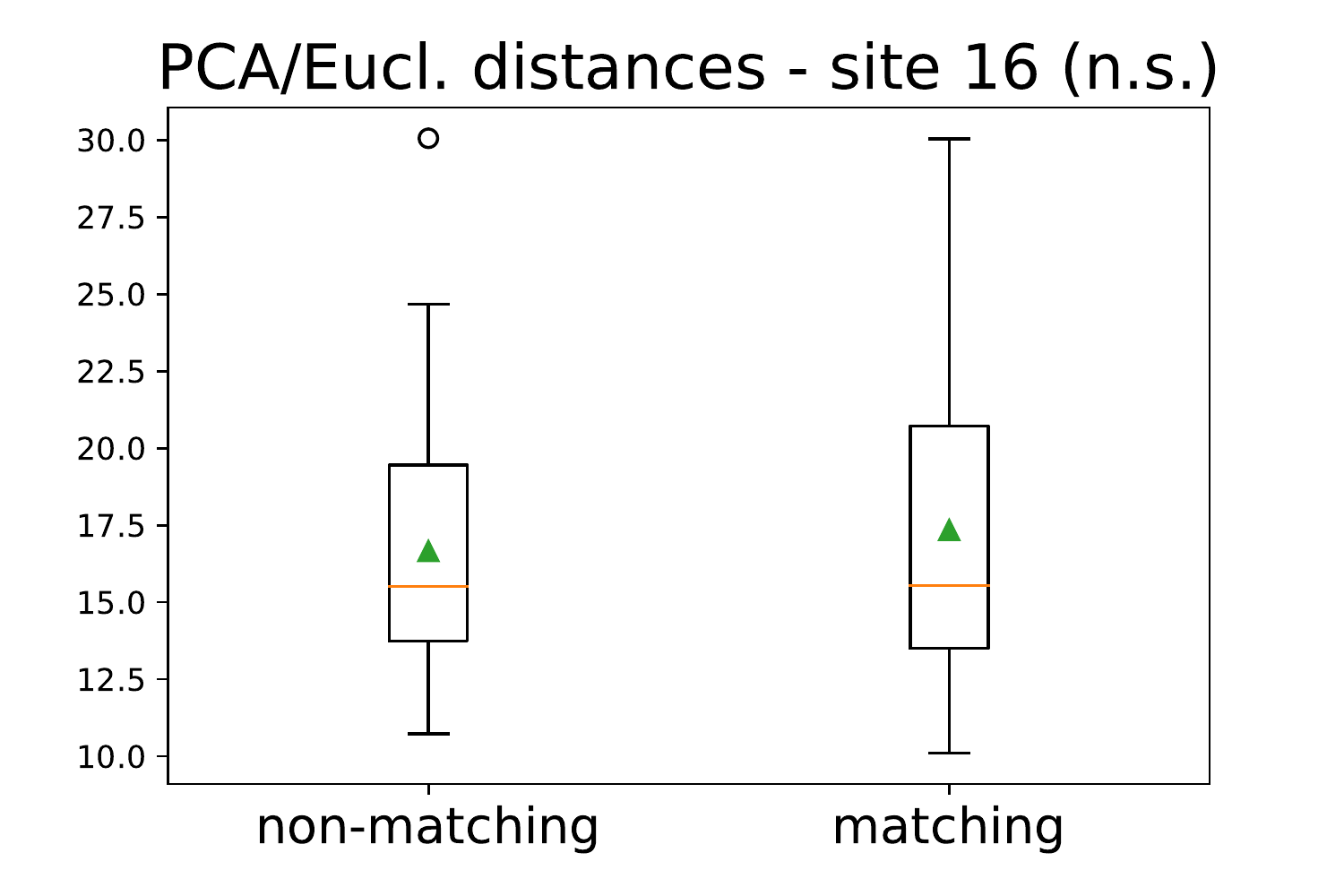}}
\subfloat{\includegraphics[width=0.25\textwidth]{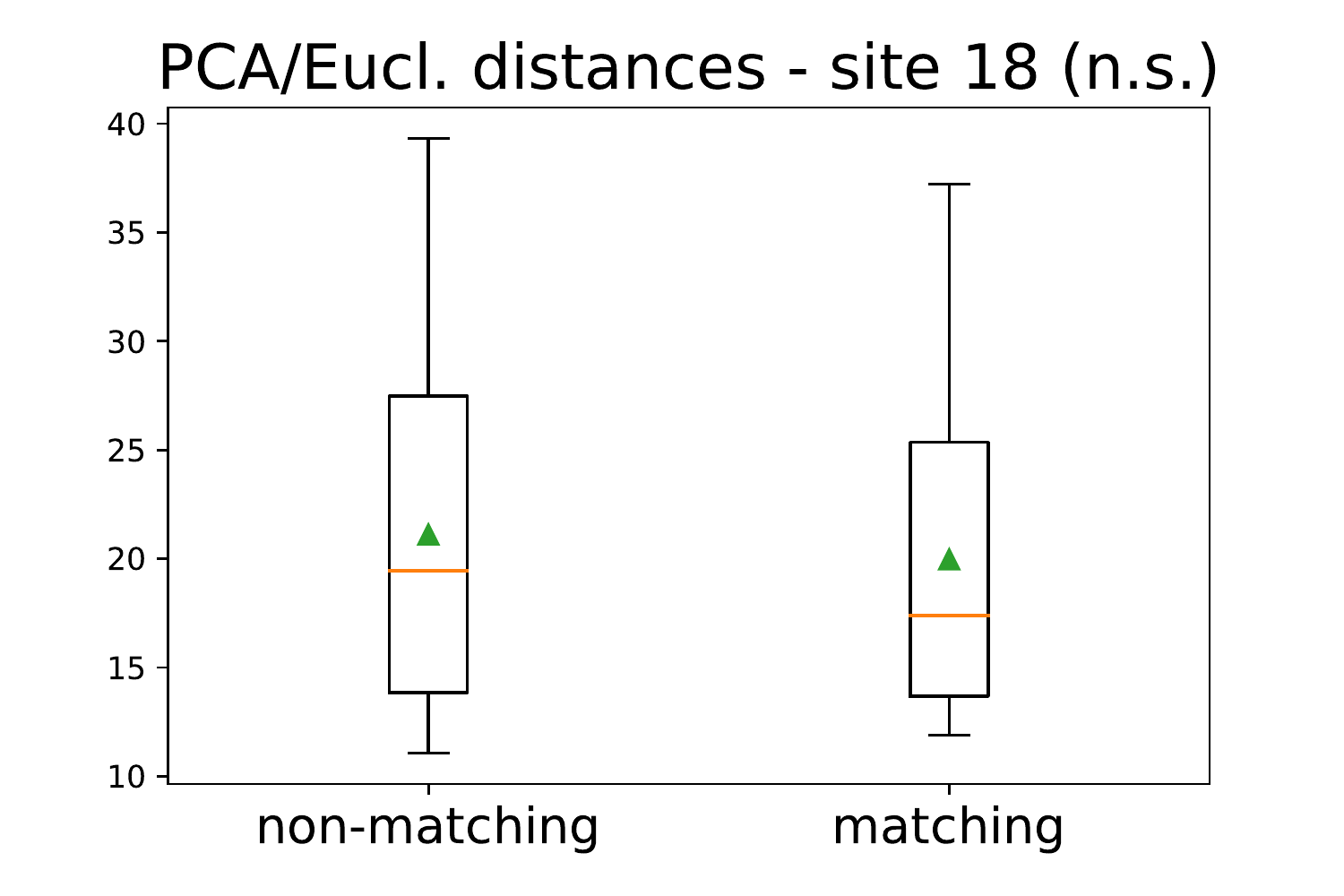}\label{subfig:site_18}}\\
\subfloat{\includegraphics[width=0.25\textwidth]{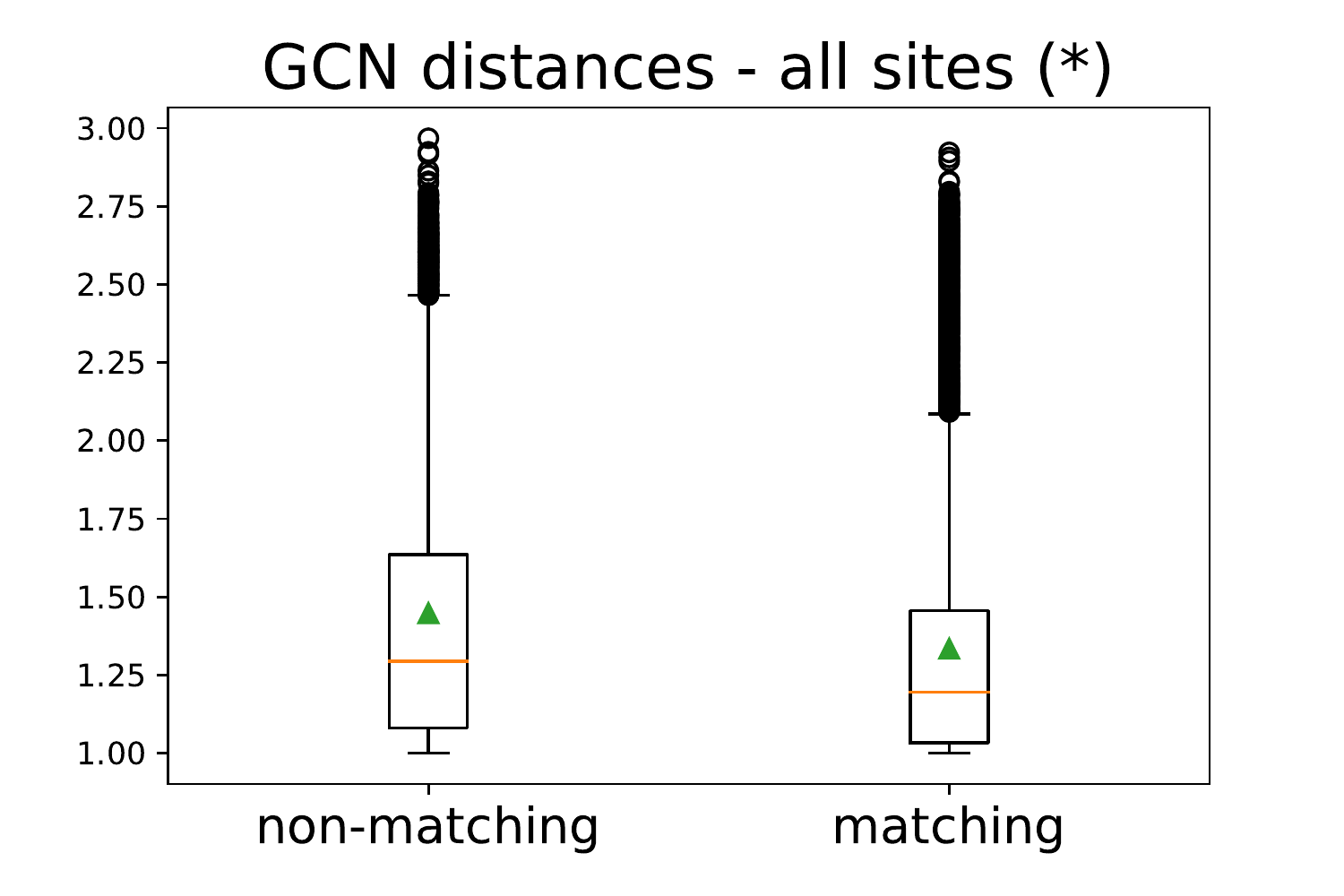}}
\subfloat{\includegraphics[width=0.25\textwidth]{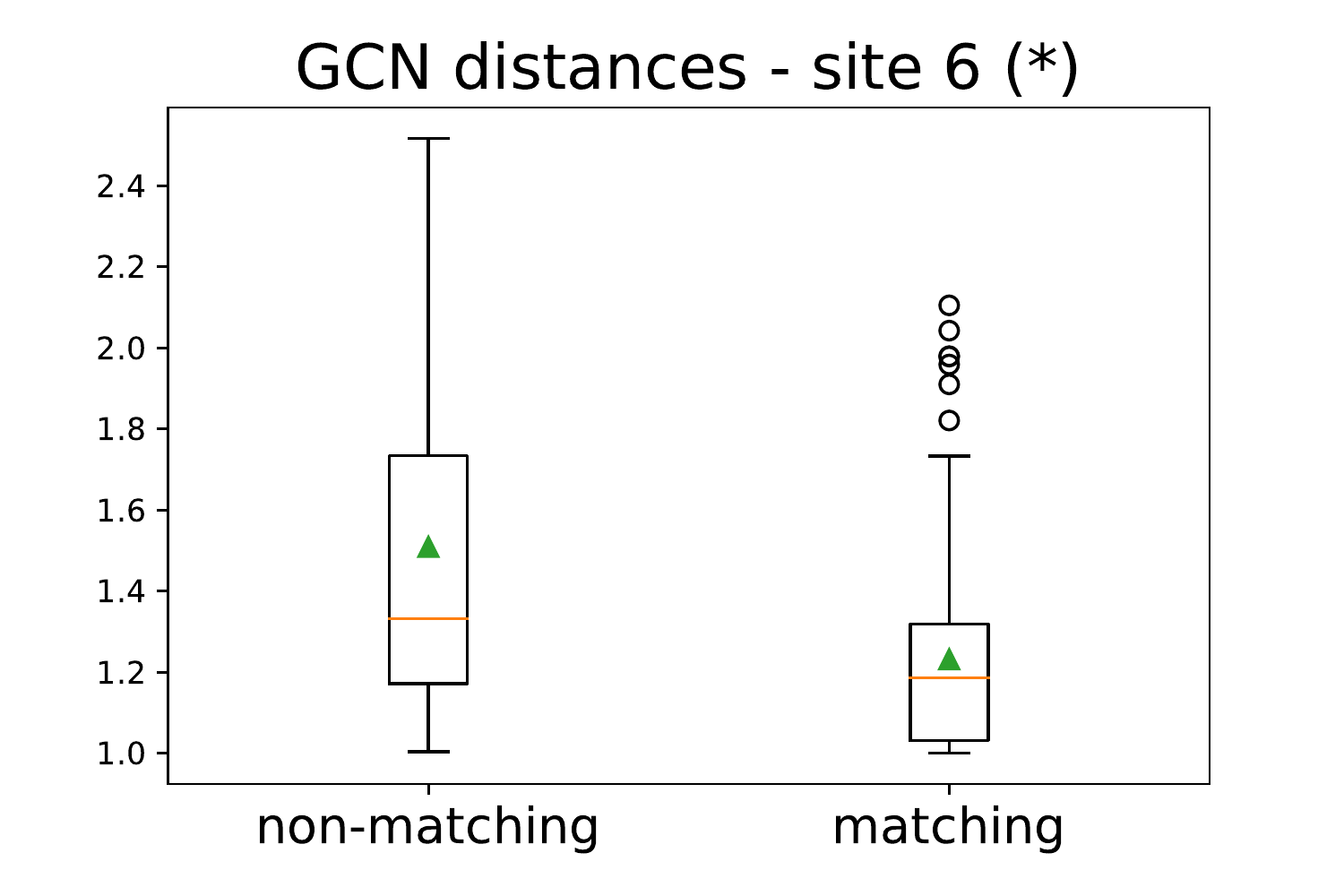}}
\subfloat{\includegraphics[width=0.25\textwidth]{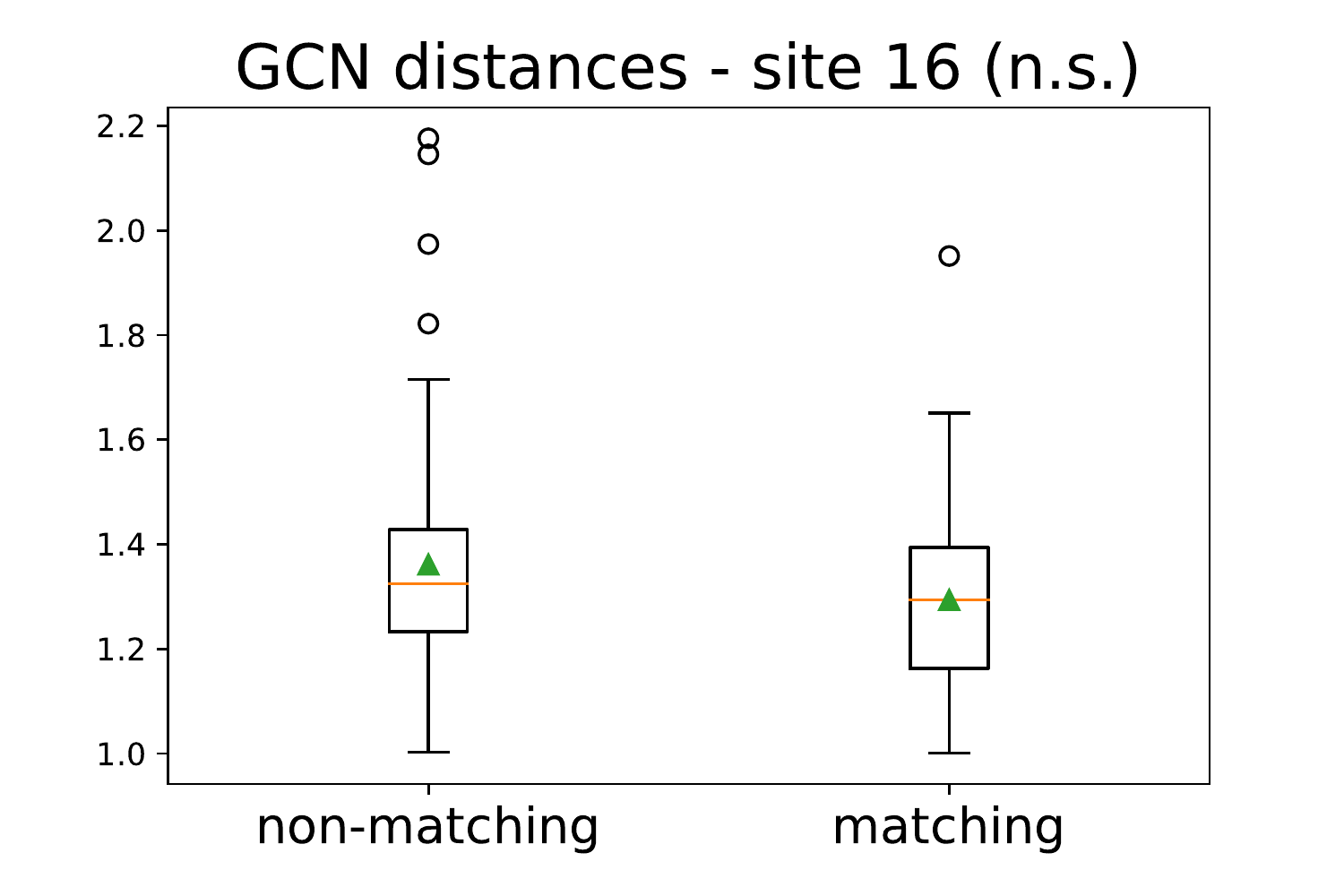}}
\subfloat{\includegraphics[width=0.25\textwidth]{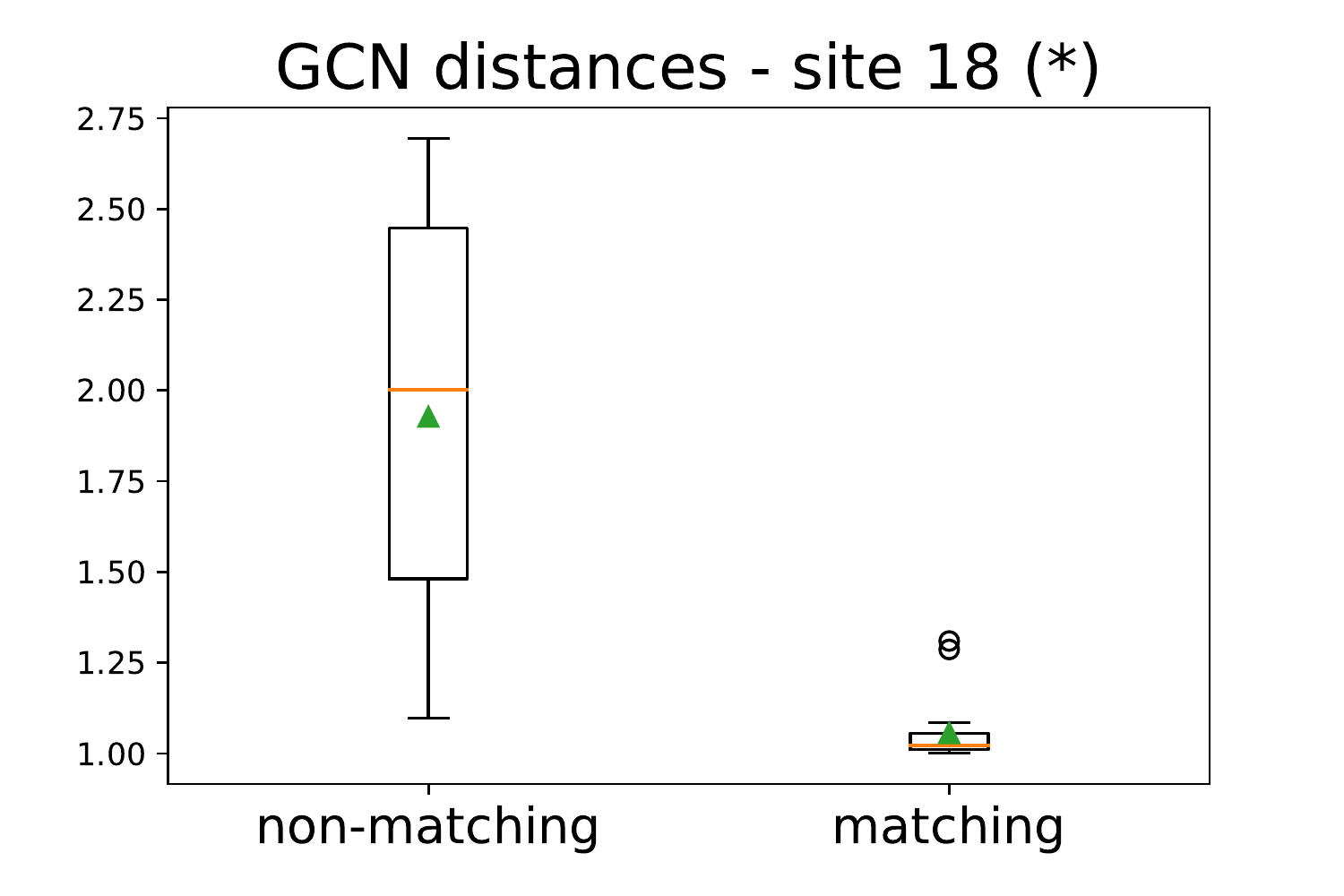}}
\caption{Box-plots showing Euclidean distances after PCA (top) and distances learned with the proposed GCN model (bottom) between matching and non-matching graph pairs of the test set. Differences between the distance distributions of the two classes (matching vs. non-matching) are indicated as significant (*) or non significant (n.s.) using a permutation test with 10000 permutations.}
\label{fig:boxplots_eucl}
\end{figure}

To illustrate how challenging the problem under consideration is, we show the pairwise Euclidean distances between functional connectivity matrices for 3 of the largest acquisition sites and the full test set after applying dimensionality reduction (PCA) in Fig.~\ref{fig:boxplots_eucl}. It can be observed that networks are hardly comparable using standard distance functions, even within the same acquisition site. ``All sites'' refers to all pairs from the test set, even if the subjects were scanned at different sites. It can be seen that the learned metric, which corresponds to the network output and is shown at the bottom of Fig.~\ref{fig:boxplots_eucl}, is significantly improving the separation between matching and non-matching pairs for the total test set, as well as for most individual sites. In order to demonstrate the learned metric's ability to facilitate a subject classification task (ASD vs control), we use a simple $k$-nn classifier with $k=3$ based the estimated distances,  and summarise results in Table~\ref{tab:classification}. Improvement in classification scores reaches 11.9\% on the total test set and up to 40\% for individual sites. Results for smaller sites are omitted, since they have very few subjects in the test set to draw conclusions from.

\begin{table}[t]
\centering
\caption{$k$-nn classification results with $k=3$ using the proposed metric and Euclidean distance following PCA.}
\label{my-label}
\begin{tabular}{@{}l|c|c|c|c|c|c@{}}
\toprule
Classification & all sites & site 6 & site 9 & site 14 & site 16 & site 18 \\ \midrule
PCA/Euclidean             & 51.0\%    & 59.3\% & 25.0\% & 30.0\%  & 64.3\%  & 50.0\%  \\ 
GCN              & \textbf{62.9\%}    & \textbf{81.5\%} & \textbf{62.5\%} & \textbf{70.0\%}  & 50.0\%  & \textbf{90.0\%}  \\ \bottomrule
\end{tabular}
\label{tab:classification}
\end{table}

\begin{figure}[t]
\centering
\subfloat[]{\includegraphics[width=0.33\textwidth]{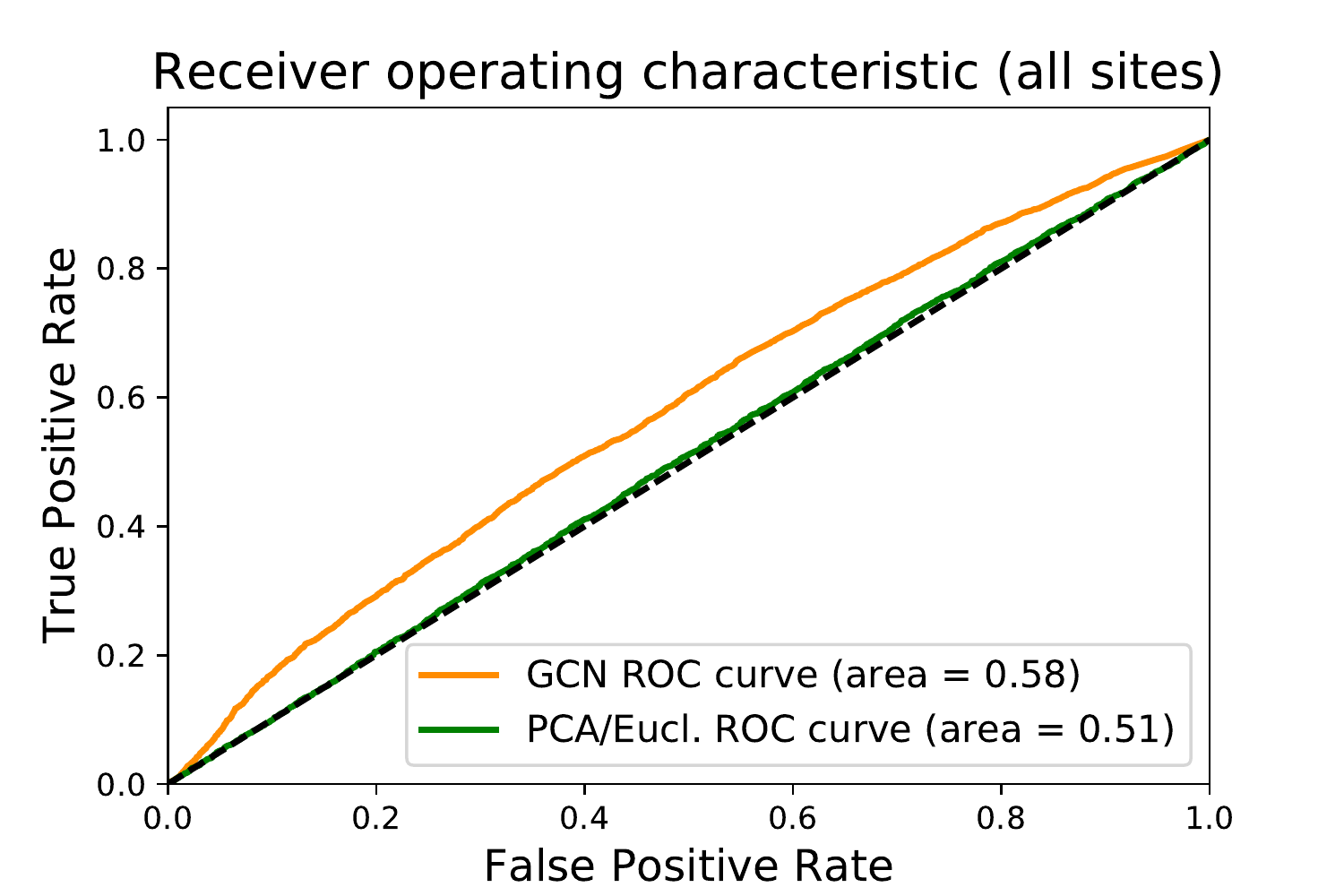}\label{subfig:roc_all_sites}}
\subfloat[]{\includegraphics[width=0.33\textwidth]{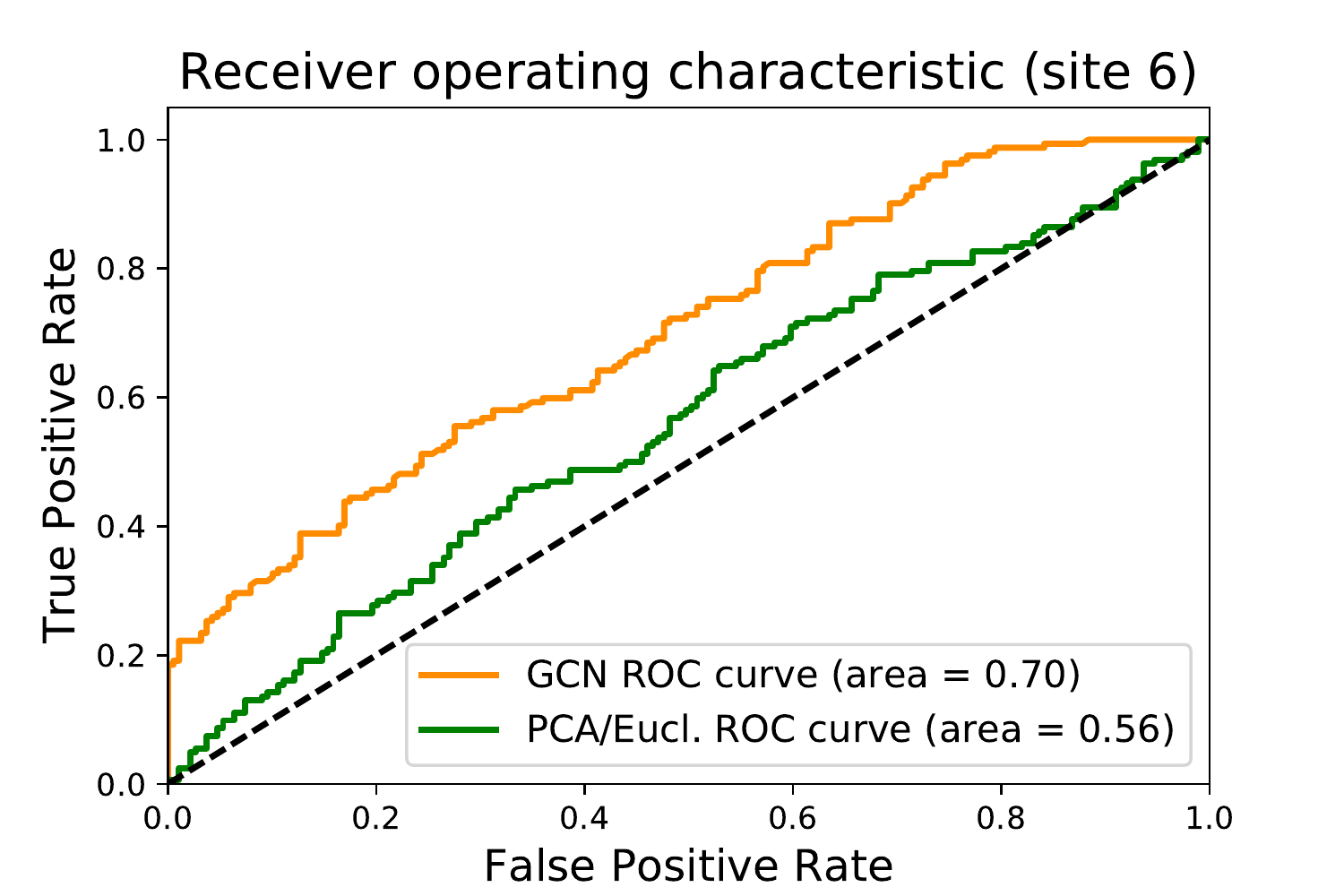}\label{subfig:site_6}}
\subfloat[]{\includegraphics[width=0.33\textwidth]{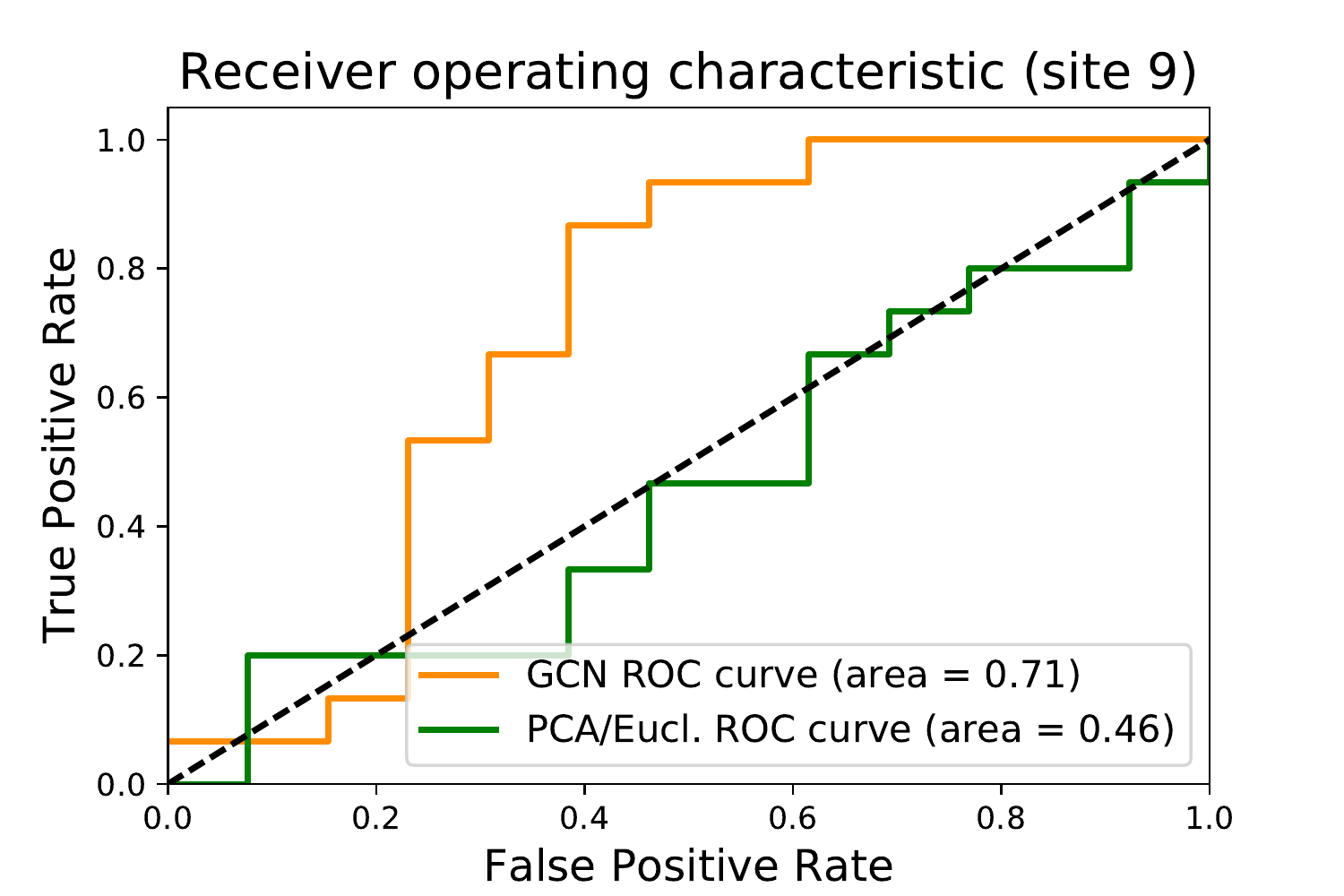}\label{subfig:site_9}}\\
\vspace{-0.3cm}
\subfloat[]{\includegraphics[width=0.33\textwidth]{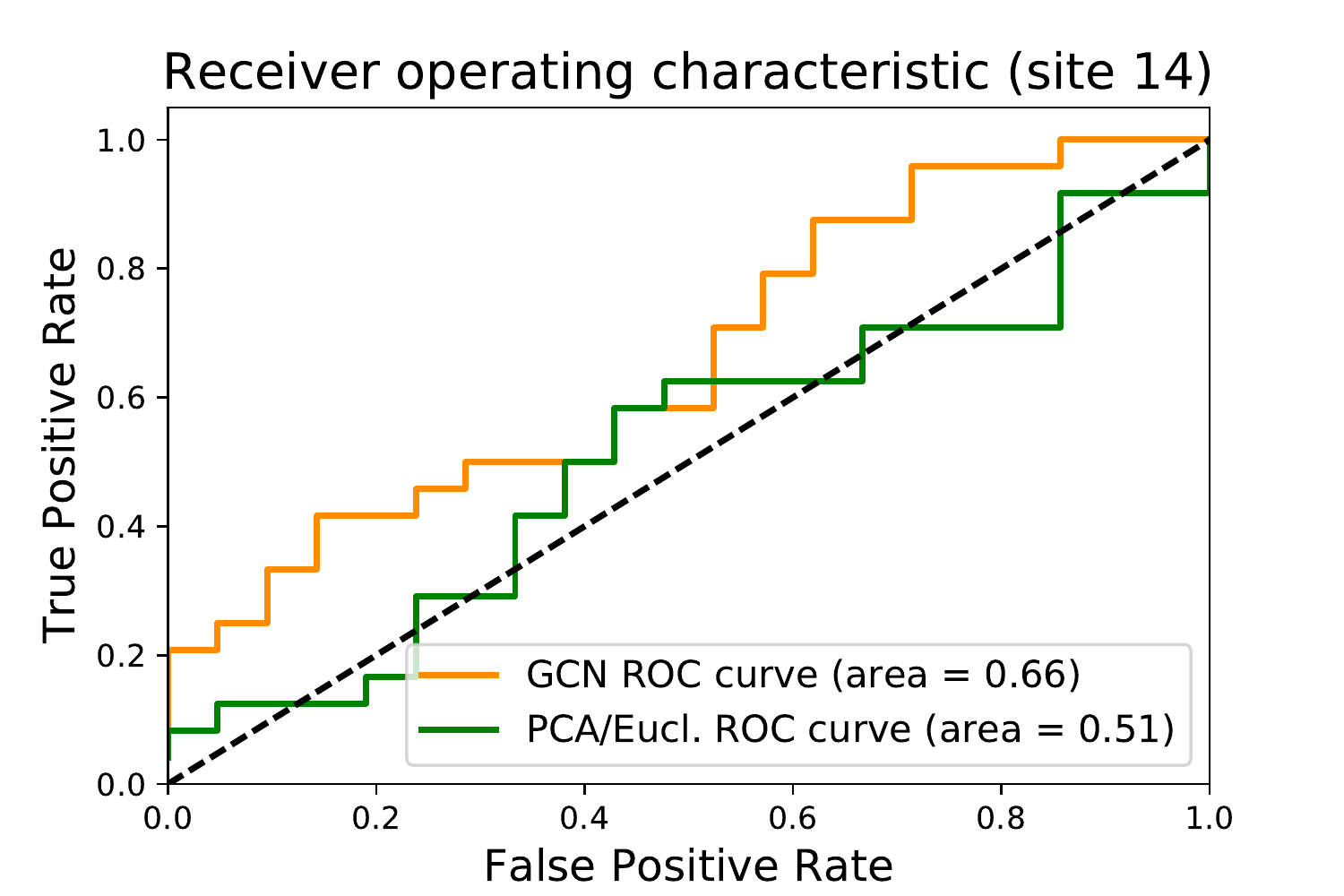}\label{subfig:site_14}} 
\subfloat[]{\includegraphics[width=0.33\textwidth]{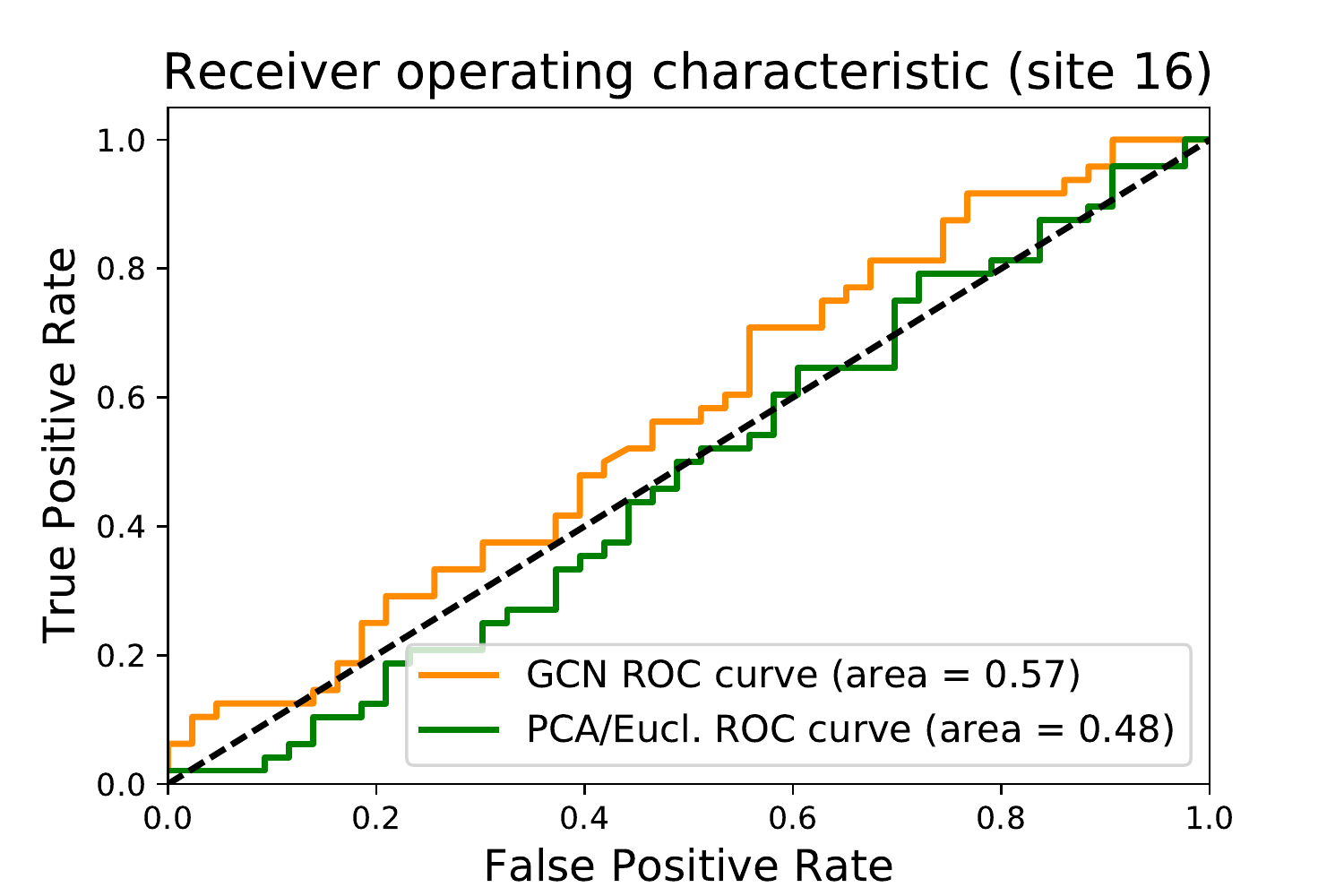}\label{subfig:site_16}}
\subfloat[]{\includegraphics[width=0.33\textwidth]{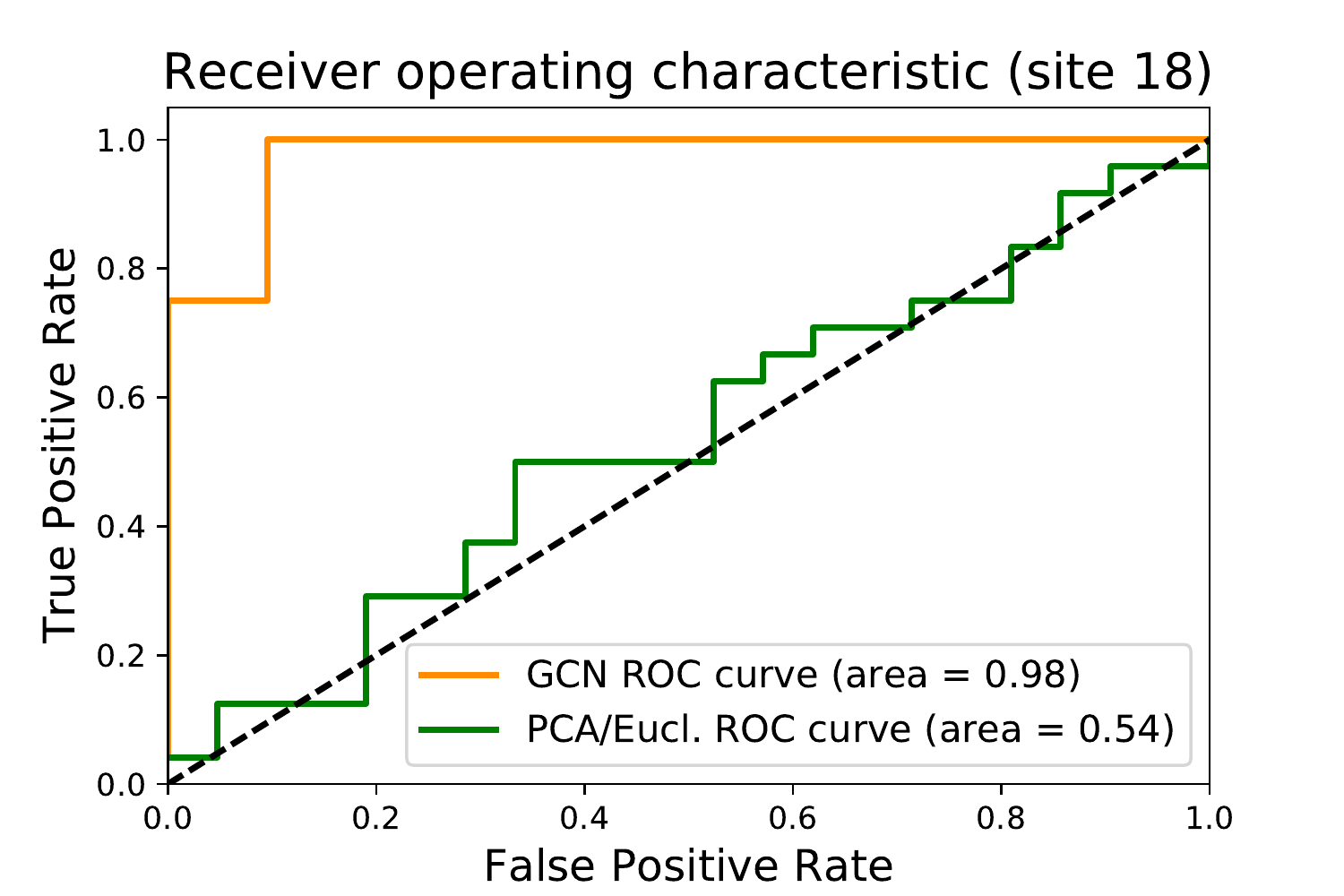}\label{subfig:site_18}}
\vspace{-0.3cm}
\caption{ROC curves and area under curve (AUC) for the classification of matching vs. non-matching graphs on the test set (a) for all sites and the 5 biggest sites (b-f) for the proposed metric and Euclidean distance.}
\label{fig:roc_curves}
\end{figure}

Fig.~\ref{fig:roc_curves} illustrates the results on the test set through receiver operating characteristic (ROC) curves for the task of classification between matching and non-matching graphs for the biggest 5 sites, as well as across all sites, along with the estimated area under curve (AUC). Fig.~\ref{subfig:roc_all_sites} shows promising results, with an overall improved performance of the proposed learned metric compared to a traditional distance measure on the whole database. The performance of the network is more striking between pairs from the same site. We obtain higher AUC values for all of the 5 biggest sites, with increases of up to 0.44 (for site 18). The limited performance for ``all sites'' could be attributed to the heterogeneity of the data across sites, as illustrated in Fig.~\ref{fig:boxplots_eucl}.

\section{Discussion}

In this work, we propose a novel metric learning method to estimate similarity between irregular graphs. We leverage the recent concept of graph convolutions through a siamese architecture and employ a loss function tailored for our task. We apply the proposed model to functional brain connectivity graphs from the ABIDE database, aiming to separate subjects from the same class and subjects from different classes. We obtain promising results across all sites, with significant increases in performance between same site pairs. While applied to brain networks, our proposed method is flexible and general enough to be applied to any problem involving comparisons between graphs, e.g. shape analysis. 

The proposed model could benefit from several extensions. The architecture of our network is relatively simple, and further improvement in performance could be obtained by exploring more sophisticated networks. A particularly exciting prospect would be to use autoencoders and adversarial training to learn lower dimensional representation of connectivity networks that are site independent. 
Additionally, exploring the use of generalisable GCNs defined in the graph spatial domain \cite{monti2016geometric} would allow to train similarity metrics between graphs of different structures. 


\bibliographystyle{splncs03}
\bibliography{references}

\begin{thebibliography}{10}
\providecommand{\url}[1]{\texttt{#1}}
\providecommand{\urlprefix}{URL }

\bibitem{abraham2016deriving}
Abraham, A., Milham, M., Di~Martino, A., Craddock, R.C., Samaras, D., Thirion,
  B., Varoquaux, G.: Deriving reproducible biomarkers from multi-site
  resting-state data: An autism-based example. NeuroImage  (2016)

\bibitem{craddock2013}
Craddock, C., Sikka, S., Cheung, B., Khanuja, R., Ghosh, S., et~al.: Towards
  automated analysis of connectomes: The configurable pipeline for the analysis
  of connectomes {(C-PAC)}. Front Neuroinform  42 (2013)

\bibitem{defferrard2016convolutional}
Defferrard, M., Bresson, X., Vandergheynst, P.: Convolutional neural networks
  on graphs with fast localized spectral filtering. In: NIPS. pp. 3837--3845
  (2016)

\bibitem{desikan2006automated}
Desikan, R.S., S{\'e}gonne, F., Fischl, B., Quinn, B.T., Dickerson, B.C.,
  et~al.: An automated labeling system for subdividing the human cerebral
  cortex on {MRI} scans into gyral based regions of interest. NeuroImage
  31(3),  968--980 (2006)

\bibitem{di2014autism}
Di~Martino, A., Yan, C.G., Li, Q., Denio, E., Castellanos, F.X., et~al.: The
  autism brain imaging data exchange: towards a large-scale evaluation of the
  intrinsic brain architecture in autism. Molecular Psychiatry  19(6),
  659--667 (2014)

\bibitem{hammond2011wavelets}
Hammond, D.K., Vandergheynst, P., Gribonval, R.: Wavelets on graphs via
  spectral graph theory. Applied and Computational Harmonic Analysis  30(2)
  (2011)

\bibitem{kipf2016semi}
Kipf, T.N., Welling, M.: Semi-supervised classification with graph
  convolutional networks. arXiv preprint arXiv:1609.02907  (2016)

\bibitem{kumar2016learning}
Kumar, B., Carneiro, G., Reid, I., et~al.: Learning local image descriptors
  with deep siamese and triplet convolutional networks by minimising global
  loss functions. In: IEEE CVPR. pp. 5385--5394 (2016)

\bibitem{livi2013graph}
Livi, L., Rizzi, A.: The graph matching problem. Pattern Analysis and
  Applications  16(3),  253--283 (2013)

\bibitem{monti2016geometric}
Monti, F., Boscaini, D., Masci, J., Rodol{\`a}, E., Svoboda, J., Bronstein,
  M.M.: Geometric deep learning on graphs and manifolds using mixture model
  {CNNs}. arXiv preprint arXiv:1611.08402  (2016)

\bibitem{niepert2016learning}
Niepert, M., Ahmed, M., Kutzkov, K.: Learning convolutional neural networks for
  graphs. In: ICML (2016)

\bibitem{raj2010network}
Raj, A., Mueller, S.G., Young, K., Laxer, K.D., Weiner, M.: Network-level
  analysis of cortical thickness of the epileptic brain. NeuroImage  52(4),
  1302--1313 (2010)

\bibitem{shervashidze2009efficient}
Shervashidze, N., Vishwanathan, S., Petri, T., Mehlhorn, K., Borgwardt, K.M.:
  Efficient graphlet kernels for large graph comparison. In: AISTATS. vol.~5,
  pp. 488--495 (2009)

\bibitem{shuman2013emerging}
Shuman, D.I., Narang, S.K., Frossard, P., Ortega, A., Vandergheynst, P.: The
  emerging field of signal processing on graphs: Extending high-dimensional
  data analysis to networks and other irregular domains. IEEE Signal Processing
  Magazine  30(3),  83--98 (2013)

\bibitem{takerkart2014graph}
Takerkart, S., Auzias, G., Thirion, B., Ralaivola, L.: Graph-based
  inter-subject pattern analysis of {fMRI} data. PloS one  9(8),  e104586
  (2014)

\bibitem{zagoruyko2015learning}
Zagoruyko, S., Komodakis, N.: Learning to compare image patches via
  convolutional neural networks. In: IEEE CVPR. pp. 4353--4361 (2015)

\end{thebibliography}

\end{document}